\pdfoutput=1                          
\documentclass[11pt]{article}
\usepackage[margin=1in,headheight=14pt]{geometry}
\usepackage{amsmath,amssymb}
\usepackage{mathpazo}                 
\usepackage{booktabs}
\usepackage{graphicx}
\usepackage{microtype}
\usepackage{xcolor}
\usepackage{fancyhdr}
\usepackage[numbers,sort&compress]{natbib}
\usepackage{caption}
\definecolor{accent}{HTML}{15487A}
\definecolor{rulegrey}{HTML}{C8C7C2}

\newif\ifpaperlinks
\IfFileExists{letltxmacro.sty}{\paperlinkstrue}{\paperlinksfalse}
\ifpaperlinks
  \usepackage{hyperref}
  \hypersetup{colorlinks=true, linkcolor=accent, citecolor=accent, urlcolor=accent}
\else
  \providecommand{\url}[1]{\texttt{#1}}
\fi

\newcommand{\apiclaudeopusfiveCost}{10.636}

\newcommand{\apiclaudeopusfiveMin}{155.2}
\newcommand{\apiclaudeopusfiveMs}{8758}

\newcommand{\apigptfiveTok}{212}

\newcommand{\apigptfivesixTok}{21}

\newcommand{\branchBestAcc}{0.968}

\newcommand{\branchWorstAcc}{0.837}

\newcommand{\dmsOursBranch}{0.818}

\newcommand{\dmsOursMathAcc}{0.560}

\newcommand{\famDistanceBand}{0.668}
\newcommand{\famFirstMove}{0.508}
\newcommand{\famReachableWithin}{0.546}

\newcommand{\frozenGapN}{12}
\newcommand{\frozenGapPP}{18}
\newcommand{\frozenHostAcc}{0.765}
\newcommand{\frozenHostBrier}{0.133}

\newcommand{\frozenN}{68}

\newcommand{\frozenOursAcc}{0.941}
\newcommand{\frozenOursBrier}{0.042}

\newcommand{\frozenSaturated}{3}

\newcommand{\frozenStates}{17}

\newcommand{\hoDistanceParity}{0.515}
\newcommand{\hoDistanceParityChance}{0.500}
\newcommand{\hoPlanProgress}{0.276}
\newcommand{\hoPlanProgressChance}{0.333}
\newcommand{\hoReachTwo}{0.570}

\newcommand{\hoSameLine}{0.980}
\newcommand{\hoSameLineChance}{0.333}
\newcommand{\jevAcc}{0.765}
\newcommand{\jevBenchAcc}{0.803}

\newcommand{\jevBrier}{0.133}
\newcommand{\ladderBudget}{4096}

\newcommand{\ladderHostedN}{10}
\newcommand{\ladderLiftN}{4}
\newcommand{\ladderLiftName}{\texttt{claude-sonnet-5}}
\newcommand{\ladderLiftPart}{0.979}
\newcommand{\ladderLiftThreshold}{0.2}
\newcommand{\ladderLiftWhole}{0.104}

\newcommand{\ladderQuestions}{480}
\newcommand{\ladderSilentGlobal}{19}
\newcommand{\ladderSilentLocal}{0}
\newcommand{\ladderStateTokens}{20,000}
\newcommand{\oursMsPerDecision}{30.9}
\newcommand{\oursPassCost}{0.000217}
\newcommand{\oursPassSeconds}{32}
\newcommand{\oursPerSecond}{32}
\newcommand{\paramCount}{1.88B parameters}

\newcommand{\paramCountShort}{1.88B}

\newcommand{\probeCond}{0.513}
\newcommand{\probeN}{300}

\newcommand{\searchFamAcc}{0.574}
\newcommand{\searchFamN}{1500}
\newcommand{\searchFamShare}{21}

\newcommand{\simCeilEvent}{0.746}

\newcommand{\simHostedN}{9}

\newcommand{\simOursAccEvent}{0.750}

\newcommand{\spatialChance}{0.343}
\newcommand{\spatialFamilies}{15}

\newcommand{\spatialN}{2,250}

\newcommand{\spatialOursAcc}{0.839}

\newcommand{\spatialOursBefore}{0.409}
\newcommand{\spatialOursGain}{2.05}

\newcommand{\spatialOursSubset}{0.844}

\newcommand{\spatialStates}{6,525}

\newcommand{\spatialTotalN}{7,305}

\newcommand{\suitesTotal}{42}
\newcommand{\surfAscii}{0.779}
\newcommand{\surfAsciiBefore}{0.457}
\newcommand{\surfAsciiChance}{0.388}
\newcommand{\surfJson}{0.936}
\newcommand{\surfJsonBefore}{0.373}
\newcommand{\surfJsonChance}{0.394}
\newcommand{\surfRows}{0.942}
\newcommand{\surfRowsBefore}{0.420}
\newcommand{\surfRowsChance}{0.394}

\newcommand{\trainFingerprints}{97,133}
\newcommand{\trainQuestions}{60,200}
\newcommand{\truncBadAcc}{0.208}
\newcommand{\truncGoodAcc}{0.909}

\newcommand{\xferUntargetedN}{13}



\makeatletter
\renewcommand\section{\@startsection{section}{1}{\z@}%
  {-3.2ex \@plus -1ex \@minus -.2ex}{2.0ex \@plus.2ex}%
  {\normalfont\large\bfseries\color{accent}}}
\renewcommand\subsection{\@startsection{subsection}{2}{\z@}%
  {-2.6ex \@plus -1ex \@minus -.2ex}{1.4ex \@plus.2ex}%
  {\normalfont\normalsize\bfseries\color{accent!85!black}}}
\makeatother

\newenvironment{tabnotes}[1][\textwidth]
  {\par\addvspace{5pt}\begin{minipage}{#1}\footnotesize\setlength{\parindent}{0pt}\raggedright}
  {\end{minipage}\par}

\newcommand{\nodataplain}{\textcolor{rulegrey!45!black}{---}}
\newcommand{\nodata}{\multicolumn{1}{c}{\nodataplain}}

\renewcommand{\headrulewidth}{0.4pt}
\renewcommand{\headrule}{\hbox to\headwidth{\color{rulegrey}\leaders\hrule height \headrulewidth\hfill}}
\fancypagestyle{plain}{\fancyhf{}\renewcommand{\headrulewidth}{0pt}\fancyfoot[C]{\footnotesize\thepage}}

\newcommand{\maybelogo}[2]{%
  \IfFileExists{#1.pdf}{\includegraphics[height=#2]{#1.pdf}}{%
  \IfFileExists{#1.png}{\includegraphics[height=#2]{#1.png}}{%
  \IfFileExists{#1.jpg}{\includegraphics[height=#2]{#1.jpg}}{}}}}

\newcommand{\Ours}{\texttt{this-that-model-1.0}}
\newcommand{\qL}{\mathrm{q}L_2}

\newcommand{\1}{\mathbf{1}}

\begin{document}
\thispagestyle{plain}

\noindent
\raisebox{-0.5\height}{\maybelogo{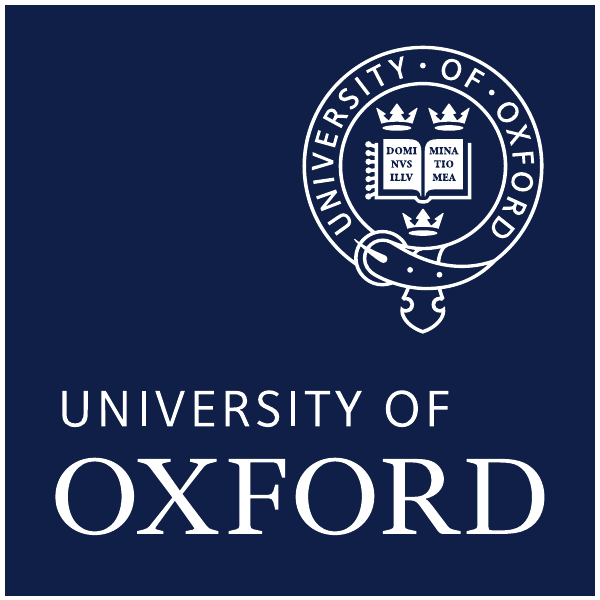}{19mm}}%
\hfill
\raisebox{-0.5\height}{\maybelogo{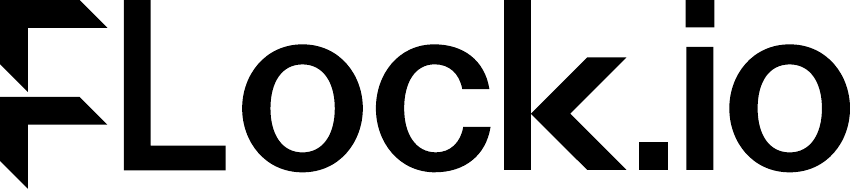}{7.5mm}}%
\par
\vspace{2.6em}

\begingroup
\centering
{\LARGE\bfseries\color{accent} \Ours\par}
\vspace{0.45em}
{\Large\bfseries A typed decision model that decides in 30\,ms, for a millionth of a cent\par}
\vspace{1.0em}
{\large Zehua Cheng\textsuperscript{1,2} \quad Wei Dai\textsuperscript{2} \quad Jiahao Sun\textsuperscript{2}\par}
\vspace{0.5em}
{\small\textsuperscript{1}University of Oxford \qquad \textsuperscript{2}FLock.io\par}
\vspace{0.25em}
{\small\texttt{ai@flock.io}\par}
\vspace{0.9em}
{\color{rulegrey}\rule{0.86\textwidth}{0.6pt}}\par
\vspace{0.6em}
\endgroup

\begin{abstract}
Software delegates more of its branches to models every year: which queue a ticket enters, whether a
command is safe to run, whether a claim clears without a person. What the program needs back is not
prose. It is one of $n$ declared options and a number it can threshold. Today that costs a round trip
to a frontier model --- hundreds of milliseconds, a per-token bill, and a parser --- for a question
that is usually a conjunction of three clauses.

\Ours{} is a 2B-parameter typed decision model. Its answer is read directly from the hidden state at
a designated position and restricted to the option set the caller declared, so no text is generated,
nothing can be malformed, and every question in a request is answered in the \emph{same} forward
pass.

It decides in \oursMsPerDecision~ms on one laptop GPU, and generates \emph{zero} output tokens doing
it. Both halves matter. At \oursMsPerDecision~ms a decision costs about what a database index lookup
costs, so it can go inside a loop, inside a retry path, inside a validator that runs on every
request --- places where a \apiclaudeopusfiveMs{}~ms API call is not an option at any price. And the
zero is not a small number, it is the absence of one: the hosted systems that answer these questions
well spend between \apigptfivesixTok{} and \apigptfiveTok{} generated tokens per question thinking
first, and are billed for every one. \Ours{} sustains \oursPerSecond{} decisions per second on one
consumer GPU and never lets the state leave the machine. This paper describes how that is built ---
the typed head, a prompt layout whose schema prefix is computed once and reused across states, a
training objective that is a proper scoring rule estimated without bias from single sampled
outcomes, and a deployment scheme in which the entire adaptation is one scalar --- and then measures
it.

On a third party's recorded cohort of \frozenN{} decision questions, on their inputs and their
wording, it scores \frozenOursAcc{} with a Brier score of \frozenOursBrier{}, against
\frozenHostAcc{} and \frozenHostBrier{} for \texttt{Jev}, the hosted service, on the same items.
One pass of our \suitesTotal{}-family internal suite takes \oursPassSeconds{} seconds and
\$\oursPassCost{} of electricity; the most accurate hosted model we measured needs
\apiclaudeopusfiveMin{} minutes and \$\apiclaudeopusfiveCost{}. We also report where it loses. On the branch-shaped families the product is for it scores
\dmsOursBranch{}, against \branchWorstAcc{} to \branchBestAcc{} for the hosted models; on
multi-step arithmetic, which a single forward pass cannot carry intermediate results through, it
scores \dmsOursMathAcc{} against their 0.98 to 1.00. And a targeted second training round improved
the five task families it was written for and transferred to none of the other \xferUntargetedN{}.
\end{abstract}

\section{The smart if-statement}

A program reaches a branch it cannot express in code. Is this refund within policy? Does this output
satisfy the instruction it was given? Is this shell command safe to run unattended? The condition is
real, it is decidable, and no regular expression will decide it.

The common answer is to call a large language model and parse what comes back. That works, and it is
expensive in a specific way. The model must produce tokens, so the cost is a decoding loop rather
than a single evaluation; the output is free-form, so the caller must parse it and handle the case
where parsing fails; and the call leaves the machine, so the state being judged leaves with it. For a
branch on a request path, the accuracy of the frontier model is rarely what hurts. The
\apiclaudeopusfiveMs{}~ms and the invoice are.

The parsing problem deserves more than a clause, because it is the part that survives into
production. A generative model asked for one of two options can return the option, the option
inside a sentence, a refusal, a markdown fence, an empty string when a reasoning budget was spent
before the answer arrived, or a plausible word that is not on the list. Each of these is a distinct
branch in the caller, and the caller must pick a behaviour for the case where none of them parse:
fail open, fail closed, or retry. That decision is a policy choice about correctness, and it is
usually made once, quickly, in an exception handler. We hit this from the measurement side while
writing this paper --- a frontier model appeared to score \truncBadAcc{} on a task it answers
correctly \truncGoodAcc{} of the time, because a truncated reply had been scored as a wrong answer
rather than as no answer --- and the general lesson is that ``the model returned nothing'' and
``the model was wrong'' are different events that a text interface makes hard to keep apart.

A \emph{typed decision model} removes the decoding loop and the parser with it. The caller declares
the answer type up front --- a choice among named options, an ordered score, or a boolean --- and
the model returns a probability distribution over exactly that set. Nothing is sampled; one forward
pass produces the hidden state at each answer position, and the distribution is a softmax over the
option labels and nothing else. The set of possible outputs is the set of declared options, so a
malformed answer is not unlikely, it is unrepresentable, and the branches above collapse into a
lookup. This paper is about building one that is small enough to run beside the application and
good enough to be trusted with the branch.

Two things follow that are not obvious from ``it returns a letter''. The first is that the number
attached to the answer is a usable probability rather than a token score. Trained against a
strictly proper scoring rule (Section~\ref{sec:objective}), the model has no way to improve its
loss except by reporting what it believes, which is what lets a caller put a threshold on the
output and route the uncertain cases somewhere else. On the one family in our measurements whose
true answer is a computed probability rather than a label, the model lands at
\simOursAccEvent{} against a ceiling of \simCeilEvent{} that no predictor can exceed, while most
hosted endpoints expose no probability at all and every one that does is worse than a constant.
The second is that $N$ questions about one state cost one pass rather than $N$, because no answer
is written back into the prompt and the decisions are therefore conditionally independent given the
input --- so a decision point that asks four things about a request is not four times the price of
one.

The number that decides whether such a thing is usable is the latency, and ours is
\oursMsPerDecision~ms. A branch that costs \oursMsPerDecision~ms can be evaluated on every request
without a budget conversation; a branch that costs \apiclaudeopusfiveMs{}~ms cannot be evaluated on
every request at all, and the usual response is to sample it, to cache aggressively, or to give up
and write a worse rule in code. One laptop GPU sustains \oursPerSecond{} decisions per second, which
is what an application tier would otherwise buy from an API for \$\apiclaudeopusfiveCost{} per
thousand questions. The model is \paramCount{}, it runs on a card that is already in the machine,
and because it runs there the state being judged never leaves --- which for a refund, a shell
command or a customer record is frequently the binding constraint rather than the price.

We should say plainly what this is not. It is not a claim to be more accurate than a frontier
model: on the harder half of our suite it is not, and on a benchmark we built and released for this
paper it is the least accurate system we measured (Section~\ref{sec:spatial}). It is a claim about
the cost of a decision, and about a shape of output that a program can rely on. The honest summary
of the trade is that a frontier model is a better judge and a worse component, and that a great
many branches in real systems need a component.

\begin{figure}[t]
\centering
\includegraphics[width=0.82\textwidth]{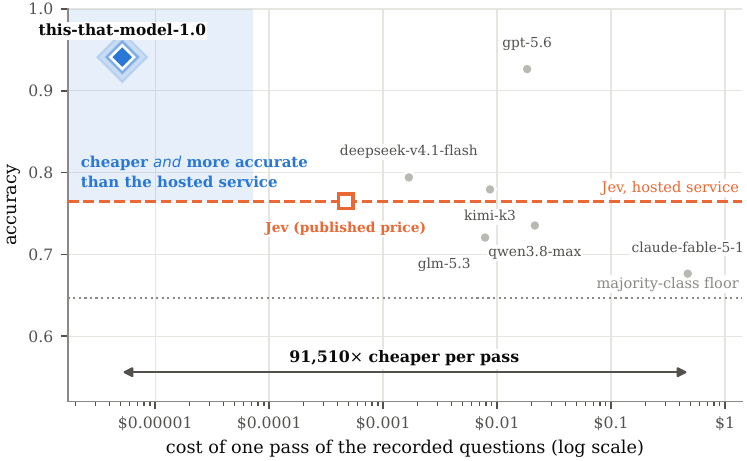}
\caption{Accuracy against what one pass of the recorded questions costs. Cost spans five orders of
magnitude; accuracy spans twenty-five points, and one system is alone in the shaded corner. Two
caveats the chart cannot carry. The hollow square is \texttt{Jev} priced at its
\emph{published} rate of \$0.042 per million input tokens with output free, applied to the token
count we measured --- its accuracy is ours to measure, its cost is theirs to state. And our own
cost is electricity at 80\,W on a card already bought, which is a different kind of number from a
price that has to cover serving and margin; the honest reading of our position against the square
is one order of magnitude, not five.}
\label{fig:frozen}
\end{figure}

\paragraph{Contributions.} (i) A typed answer head that makes malformed output impossible by
construction and answers $N$ questions about one state in a single forward pass
(Section~\ref{sec:head}). (ii) A prompt layout whose cost is amortised over states rather than paid
per question (Section~\ref{sec:layout}). (iii) A training objective that is a strictly proper
scoring rule and can be estimated without bias from a single observed outcome per question, using a
paired control variate (Section~\ref{sec:objective}). (iv) A deployment scheme in which the entire
adaptation is a single scalar, so rollback is exact rather than approximate
(Section~\ref{sec:lambda}). (v) Measurements against \texttt{Jev}, a hosted commercial service, on inputs we did
not choose, and against \simHostedN{} hosted frontier models on identical questions whose answer is
a computed \emph{distribution} rather than a sampled label (Section~\ref{sec:results}).

\section{Related work}

\paragraph{Typed and constrained model outputs.} Structured decoding --- grammars, JSON schemas,
regular-expression constraints~\cite{willard2023efficient,geng2023grammar} --- makes a generative
model's output parseable, but it remains a
decoding loop, and the constraint is applied to the tokens rather than to the hypothesis space. The
typed-decision formulation instead reads the answer from a hidden state and normalises over the
declared options, which removes the loop rather than disciplining it. The commercial ``System One''
line of models~\cite{almeida2026systemone}, named for the fast automatic mode of
reasoning~\cite{kahneman2011thinking}, popularised this interface for application developers; \texttt{decider-2b} is a
publicly released 2B typed-decision checkpoint and inference stack under Apache-2.0: the
architecture family our model belongs to, and the source of the wire format that
Section~\ref{sec:head} formalises. NanoJev~\cite{nanojev2026} is an open 0.6B reproduction of the
same interface under the MIT licence; it contributes the environment simulator and the recorded hosted-service cohort we
measure against in Section~\ref{sec:results}, and its own checkpoint appears in
Table~\ref{tab:frozen} as a baseline.

\paragraph{Proper scoring rules and calibration.} That a decision model's probability must be
usable, not merely rankable, is old: strictly proper scoring rules~\cite{gneiting2007proper}, of which the logarithmic and the
Brier score~\cite{brier1950verification} are the two canonical members, are optimised uniquely at
the true conditional distribution. Calibration is usually measured after the fact, by binning predictions against observed
frequencies~\cite{guo2017calibration,naeini2015obtaining,nixon2019measuring}, a procedure whose
estimator is itself sensitive to the binning~\cite{nixon2019measuring}. What is less common is
evaluating against a \emph{known} distribution rather than a sampled label, as simulation-based
calibration does for posterior inference~\cite{cook2006validation,talts2018validating}; Section~\ref{sec:results} does that where the environment makes it
possible, and Equation~\eqref{eq:decomp} says exactly what is lost when it is not.

\paragraph{Weight-space adaptation.} Interpolating between a base and a fine-tuned checkpoint along
their difference vector is a standard tool for trading specialisation against
generality~\cite{ilharco2023editing,wortsman2022model,yadav2023ties}, and one answer among several
to the forgetting that full fine-tuning induces~\cite{kirkpatrick2017overcoming}; parameter-efficient
adaptation~\cite{hu2022lora} is another, which we did not use because the budget allowed the full
fine-tune. We use it
for an operational property rather than for accuracy --- the zero setting is an exact identity, so a
rollback needs no requalification (Section~\ref{sec:lambda}).

\paragraph{Linear-attention backbones.} Delta-rule and related linear-attention
layers~\cite{katharopoulos2020transformers,yang2024gated} replace the quadratic attention matrix
with a fixed-size recurrent state, giving $O(L)$ cost in sequence length. Hybrids that interleave a
minority of full-attention layers recover most of the quality~\cite{yang2025qwen3}. This is
background for us, not contribution, and it is why a 32k-token state is affordable at all.

\section{Method}

\subsection{The typed answer head}
\label{sec:head}

Let $x$ be the rendered request: a state $S$ (a string, or any JSON value serialised compactly) and
questions $q_1,\dots,q_N$, where question $k$ declares an ordered option set
$A_k = (o_{k,1},\dots,o_{k,n_k})$. The request is rendered so that each question ends at a
designated \emph{answer slot} --- a fixed token position $s_k$ --- and the model is run once:
\begin{equation}
H \;=\; f_\theta(x) \in \mathbb{R}^{L \times d},
\qquad
h_k \;=\; H_{s_k} .
\end{equation}
Each option is assigned a label token whose embedding row in the unembedding matrix $W \in
\mathbb{R}^{|V| \times d}$ is $w_{\ell}$; write $\ell(k,j)$ for the label token of option $j$ of
question $k$. The answer distribution is the softmax over \emph{only} those rows:
\begin{equation}
p_k(j \mid x) \;=\;
\frac{\exp\!\big(\langle w_{\ell(k,j)},\, h_k\rangle / \tau\big)}
     {\sum_{j'=1}^{n_k} \exp\!\big(\langle w_{\ell(k,j')},\, h_k\rangle / \tau\big)},
\qquad j = 1,\dots,n_k .
\label{eq:head}
\end{equation}
Three properties follow directly from~\eqref{eq:head} and are the reason for the design.

\emph{The output cannot be malformed.} The support of $p_k$ is $A_k$ by construction. There is no
parser and no retry path, because there is no string. A generative model can be constrained towards
this behaviour by decoding rules, but the probability it assigns to the declared set is
$\sum_{j} P(\text{option } j) < 1$ in general, and the residual mass has to be handled somewhere.
Here it is identically zero.

\emph{The cost is one evaluation, not a loop.} Generating $T$ answer tokens costs $T$ sequential
forward passes and $T$ round trips through the sampler. Equation~\eqref{eq:head} costs one pass and
one inner product per option.

\emph{The $N$ answers share that pass.} No answer letter is ever written into the context, so slot
$k$ attends to the state and to all $N$ questions, but never to another answer. The model therefore
factorises the joint as
\begin{equation}
p(a_1,\dots,a_N \mid x) \;=\; \prod_{k=1}^{N} p_k(a_k \mid x),
\end{equation}
which is the conditional-independence assumption a caller wants when it asks $N$ separate questions
about one state. Answering them separately would cost $N$ passes over $|S|$ tokens; answering them
jointly costs one pass over $|S| + \sum_k |q_k| + N$.

Labels are chosen so that every option costs exactly one token. For $n_k \le 10$ the rendering is
the familiar \texttt{(A)}\dots\texttt{(J)}; beyond that we use a table of $255$ upper-case strings
that the tokenizer maps to single ids, which keeps~\eqref{eq:head} a single gather regardless of how
many options the caller declares.

\subsection{Rendering: where the state goes}
\label{sec:layout}

The obvious layout puts the state first and the questions after it. It has a defect at scale. In
production the same \emph{schema} --- the questions and their option sets --- is asked about many
different states: one classifier, a million tickets. With the state first, nothing is shared between
requests.

We therefore also support a schema-first layout, in which the token stream is
\begin{equation}
\underbrace{\big[\,q_1, A_1, \dots, q_N, A_N\,\big]}_{\text{prefix } P,\ \text{depends only on the schema}}
\;\Vert\;
\underbrace{\big[\,S\,\big]}_{\text{the state}}
\;\Vert\;
\underbrace{\big[\,\text{slot}_1, \dots, \text{slot}_N\,\big]}_{\text{answer slots}} .
\end{equation}
The prefix is byte-identical for every state, so its attention keys and values --- and, in the
linear-attention layers, its recurrent state --- are computed once and reused, which is the same
sharing that prefix caching exploits in a serving stack~\cite{kwon2023efficient}. Serving $m$ states
under one schema costs
\begin{equation}
C_{\text{schema-first}} \;=\; |P| \;+\; m\,(|S| + N)
\qquad\text{instead of}\qquad
C_{\text{state-first}} \;=\; m\,(|P| + |S| + N),
\end{equation}
a saving of $(m-1)|P|$ tokens of prefill. Schemas with many options are exactly the case where $|P|$
is large, so the saving grows with the thing that would otherwise hurt. Training mixes the two
layouts so that one checkpoint serves both.

\begin{figure}[t]
\centering
\includegraphics[width=0.82\textwidth]{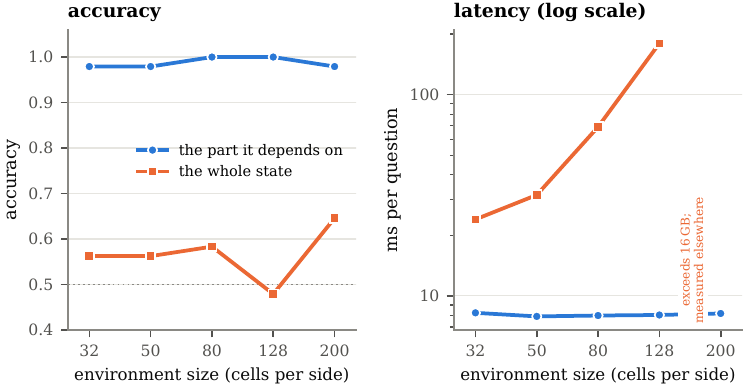}
\caption{Two panels rather than two y-axes. Reading the part the answer depends on is flat in both
accuracy and latency as the world grows; reading the whole map is worse at every size and costs an
order of magnitude in time by $128\times128$. The whole-state input at $200\times200$ is
\ladderStateTokens{} tokens, which does not fit the 16\,GB card the latency panel was measured on;
that rung's accuracy was measured later on an 80\,GB card and its latency is therefore absent
rather than merged in from another machine.}
\label{fig:ladder}
\end{figure}

\paragraph{Why long states are affordable.} The backbone is a hybrid: most layers are
delta-rule linear-attention layers, which carry a fixed-size matrix state $S_t$ updated per token,
\begin{equation}
S_t \;=\; S_{t-1}\big(I - \beta_t\, k_t k_t^{\!\top}\big) \;+\; \beta_t\, v_t k_t^{\!\top},
\qquad
o_t \;=\; S_t\, q_t ,
\end{equation}
interleaved with a minority of full-attention layers. The linear layers cost $O(L)$ in sequence
length rather than $O(L^2)$ and keep a constant amount of state per token, which is what makes a
32k-token state a routine request rather than an incident. We did not design this backbone; we
mention it because it is the reason the numbers in Figure~\ref{fig:ladder} are shaped the way they
are.

\subsection{Training against a proper scoring rule}
\label{sec:objective}

A decision model is useful only if the number it returns can be thresholded, so the training
objective must be a \emph{proper scoring rule}: one whose expected value is optimised, uniquely, by
reporting the true conditional distribution. Cross-entropy is one. The Brier score,
\begin{equation}
\mathrm{B}(p, y) \;=\; \lVert p - e_y \rVert^2 ,
\end{equation}
is another, and it is the one that matches how the output is consumed --- a squared penalty on the
probability itself rather than on its logarithm, so it does not pay unboundedly for rare confident
errors. We train on a convex combination of the two.

\paragraph{Learning from what actually happened.} Some of our supervision is not a label but an
\emph{outcome}: an action was taken in a stochastic environment and something happened. For these we
want to optimise the expected Brier reward
\begin{equation}
R(p) \;=\; \mathbb{E}_{Y}\big[\,2\,p_Y - \lVert p \rVert^2\,\big]
\qquad
\big(= -\,\mathbb{E}_Y\,\mathrm{B}(p, Y) + \text{const}\big),
\end{equation}
where $Y$ is the option the environment rewarded. Given $M$ draws $a_1,\dots,a_M \sim p$ and one
observed outcome $y$, the quantity
\begin{equation}
\widehat{R}
\;=\;
\frac{2}{M}\sum_{i=1}^{M} \1[a_i = y]
\;-\;
\frac{2}{M(M-1)}\sum_{i<j} \1[a_i = a_j]
\label{eq:rhat}
\end{equation}
is unbiased for $R(p)$: the first term estimates $2p_y$, and the second is the $U$-statistic that
estimates $\lVert p \rVert^2$ without the bias that $\frac{1}{M}\sum_i p_{a_i}$ would carry. The
gradient estimator assigns each draw its own credit $r_i$ (its terms of~\eqref{eq:rhat}) and
subtracts a paired control variate $b_i$ formed by substituting $p$ for the sampled indicators,
\begin{equation}
\nabla_\theta \widehat{R}
\;=\;
\sum_{i=1}^{M} \big(r_i - b_i\big)\,\nabla_\theta \log p(a_i),
\qquad
\mathbb{E}\big[b_i \,\nabla_\theta \log p(a_i)\big] = 0 ,
\end{equation}
so the variance reduction costs nothing in bias. In practice $M = 32$ draws per question is enough
that the estimator is quieter than the data.

\paragraph{Abstention.} A decision system that may decline is a selective
predictor~\cite{geifman2017selective,el2024leave}, and a caller that offers ``none of the above''
must not leak information by offering it. During training we insert an abstention option into a fraction of questions --- with
the correct answer removed in exactly the cases where abstaining is correct --- so that the presence
of the option is uninformative about whether it is the answer.

\subsection{Shipping the adaptation as one number}
\label{sec:lambda}

Let $\theta_0$ be the checkpoint we started from and $\theta_\star$ the checkpoint after training.
We deploy not $\theta_\star$ but the family
\begin{equation}
\theta(\lambda) \;=\; \theta_0 \;+\; \lambda\,\Delta,
\qquad \Delta \;=\; \theta_\star - \theta_0,
\qquad \lambda \in [0, 1],
\label{eq:lambda}
\end{equation}
shipped as a single scalar alongside the two checkpoints. This is worth a paragraph for an
operational reason rather than a scientific one: $\theta(0) = \theta_0$ \emph{exactly}, bit for bit,
because~\eqref{eq:lambda} is an identity and not an approximation. A rollback is therefore a
configuration change with a proof attached, not a redeploy that has to be re-qualified, and
intermediate $\lambda$ gives a graded response when a regression is found in production. We ship
$\lambda = 1$ and report the sensitivity of the headline numbers to $\lambda$ in
Section~\ref{sec:results}.

\paragraph{Fitting the work on one card.} The adaptation is a full-parameter fine-tune of a 2B model
on a single 16\,GB laptop GPU, which is only possible with all of: 8-bit optimiser moments~\cite{dettmers2022optimizers} (halving
the $2\times$ fp32 state per parameter that dominates the budget), gradient checkpointing,
token-bucketed batching that keeps the number of distinct $(B, T)$ shapes near twenty-four so
kernels compile once, and a hard cap on the allocator so that an over-large batch raises instead of
spilling into host memory. None of this is novel; all of it is load-bearing, and the absence of any
one of them turns the run into a crash.

\section{Measurements}
\label{sec:results}

\subsection{Against \texttt{Jev}, on inputs neither of us chose}

Vendor comparisons usually reduce to each side quoting its own benchmark. One exception is
available. NanoJev~\cite{nanojev2026}, an open reproduction of the typed-decision interface, put
\frozenN{} questions about local geometry across \frozenStates{} states to the hosted \emph{Jev}
service~\cite{almeida2026systemone} and published the
whole exchange: the rendered inputs, the service's replies, the ground truth, and an input hash per
item. Any system can therefore be placed on exactly those inputs, with exactly that wording, and the
comparison is auditable by anyone who fetches the recording. Ground truth is computed from the
environment rather than annotated, so the top of the scale is exact rather than nominal.
Table~\ref{tab:frozen} also carries NanoJev's own 0.6B model, which was trained on this style of
question and is the strongest small open system we are aware of on it.

\begin{table}[t]
\centering\small
\caption{The same \frozenN{} recorded questions over \frozenStates{} states, sorted by accuracy.}
\label{tab:frozen}
\begin{tabular}{l rrr r@{}l r@{}l}
\toprule
system & accuracy & Brier $\downarrow$ & NLL $\downarrow$
       & \multicolumn{2}{c}{ms per question} & \multicolumn{2}{c}{cost of one pass (\$)} \\
\midrule
majority-class baseline & 0.647 & \nodata & \nodata & \multicolumn{2}{c}{\nodataplain} & \multicolumn{2}{c}{\nodataplain} \\
\texttt{claude-fable-5-1} & 0.676 & \nodata & \nodata & 2395 &  & 0 & .471 \\
\texttt{glm-5.3} & 0.721 & 0.204 & 0.601 & 819 &  & 0 & .008 \\
\texttt{qwen3.8-max} & 0.735 & 0.263 & 2.681 & 1014 &  & 0 & .021 \\
\texttt{NanoJev-0.6B} & 0.750 & 0.166 & 0.479 & \multicolumn{2}{c}{\nodataplain} & \multicolumn{2}{c}{\nodataplain} \\
\textbf{Jev} & 0.765 & 0.133 & 0.403 & \multicolumn{2}{c}{\nodataplain} & \multicolumn{2}{c}{\nodataplain} \\
\texttt{kimi-k3} & 0.779 & 0.143 & 0.430 & 989 &  & 0 & .009 \\
\texttt{deepseek-v4.1-flash} & 0.794 & \nodata & \nodata & 808 &  & 0 & .002 \\
\texttt{gpt-5.6} & 0.926 & \nodata & \nodata & 1180 &  & 0 & .018 \\
\textbf{\Ours} & \textbf{0.941} & \textbf{0.042} & \textbf{0.126} & \textbf{30} & \textbf{.9} & \textbf{0} & \textbf{.000014} \\
 
\bottomrule
\end{tabular}
\begin{tabnotes}
A dash in the Brier and NLL columns means the endpoint returns no token probabilities, so those
quantities are not observable there --- not that they are poor. The time column is the median
\emph{per-question} latency measured at the client. We do not report the wall clock of the whole
pass, because that depends on how many requests our harness chose to keep in flight and is a
property of the harness rather than of the system. Cost is for one pass of all \frozenN{}
questions; for \Ours{} it is electricity at 80\,W and \$0.30 per kWh. A further
\frozenSaturated{} frontier models answered every question correctly and are omitted: at \frozenN{}
items they are saturated and rank nothing. \texttt{NanoJev-0.6B} is the open reproduction that made
the recording; like \texttt{Jev} it was not timed by us, so its time and cost cells are
blank rather than zero.
\end{tabnotes}
\end{table}

\subsection{Whether the probability means anything}

Accuracy cannot tell you whether a probability is honest, because a label carries no probability of
its own. In an environment with a stochastic actuator it can. Let the intended action be $a^\star$,
executed with reliability $\rho$ and otherwise replaced by a uniform draw from the remaining
actions. Then the probability that the executed move is collision-free is not estimated but derived:
\begin{equation}
q \;=\; \rho\,\1[\mathrm{safe}(a^\star)]
      \;+\; \frac{1-\rho}{|A|-1}\sum_{a\neq a^\star}\1[\mathrm{safe}(a)] .
\label{eq:q}
\end{equation}
The model trains on a single Bernoulli draw $Y \sim \mathrm{Bern}(q)$ and is scored against $q$,
which it never sees. We report $\qL = \mathbb{E}\lVert p - q\rVert^2$.

\begin{figure}[tb]
\centering
\includegraphics[width=0.82\textwidth]{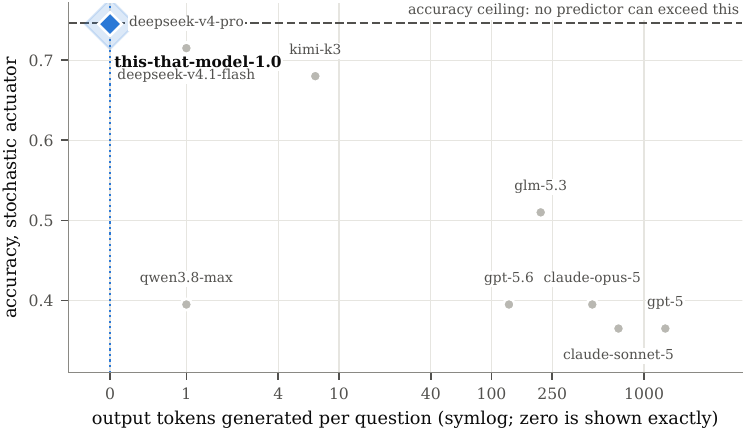}
\caption{Accuracy on the stochastic-actuator family against the number of tokens each system
generated before answering. The relationship runs the wrong way for a reader who expects
test-time computation to help: the systems that generated the most tokens are the least accurate,
and the three most accurate answer in one token or none. The environments make a noisy actuator
safe more often than not, and a model that reasons its way to caution is reasoning away from the
answer. \Ours{} sits at zero tokens by construction --- Equation~\eqref{eq:head} has no decoding
step for tokens to be spent in --- and on the accuracy ceiling.}
\label{fig:thinking}
\end{figure}

This is worth the trouble because sampled labels hide exactly this quantity. Decomposing the Brier
score against the sampled outcome,
\begin{equation}
\mathbb{E}_Y\big[(p - Y)^2\big] \;=\; (p - q)^2 \;+\; q(1-q),
\label{eq:decomp}
\end{equation}
the second term depends on the environment and not on the model. It therefore cancels in
\emph{differences} between models --- label-based comparisons are informative --- but it does not
cancel in levels, so no single Brier number on sampled data says how far a model is from the truth.
Equation~\eqref{eq:q} closes that gap. A further consequence of~\eqref{eq:decomp} is a ceiling: no
predictor can exceed $\mathbb{E}[\max(q, 1-q)]$ accuracy on such labels, however good it is.

\begin{figure}[tb]
\centering
\includegraphics[width=0.82\textwidth]{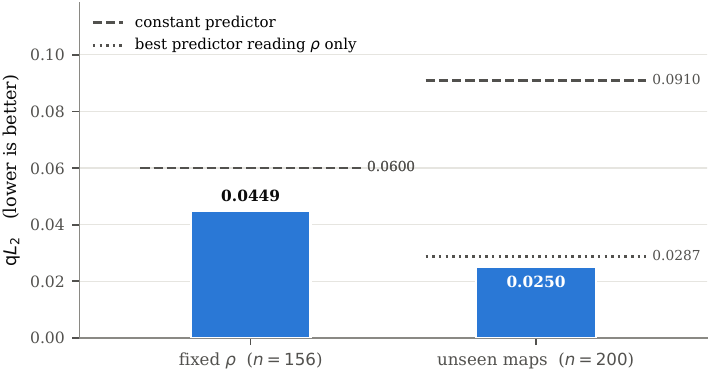}
\caption{Squared distance to the computed answer distribution, against the two predictors that
learn nothing. The bar is below both rules in both regimes; the point of drawing the rules is that
a model can look calibrated by copying $\rho$ out of its prompt, and this one is not doing that.}
\label{fig:calib}
\end{figure}

We report $\qL$ beside two predictors that learn nothing: a constant equal to $\mathbb{E}[q]$, and
the best predictor that reads $\rho$ out of the prompt and ignores the environment entirely. The
second is the one that matters --- a model can look well calibrated while having learned only to
copy a number out of its input (Table~\ref{tab:calib}, Figure~\ref{fig:calib}).

\begin{table}[t]
\centering\small
\caption{\simHostedN{} hosted systems and \Ours{} on identical questions.}
\label{tab:calib}
\begin{tabular}{l rr rr rr}
\toprule
 & \multicolumn{2}{c}{stochastic actuator} & \multicolumn{2}{c}{local geometry} & no & cost of \\
\cmidrule(lr){2-3} \cmidrule(lr){4-5}
system & accuracy & $\qL\downarrow$ & accuracy & $\qL\downarrow$ & answer & one pass (\$) \\
\midrule
accuracy ceiling (computed) & 0.746 & \nodata & 1.000 & \nodata & \nodata & \nodata \\
always-yes baseline & 0.703 & \nodata & \nodata & \nodata & \nodata & \nodata \\
constant predictor & \nodata & 0.0962 & \nodata & 0.4934 & \nodata & \nodata \\
\midrule
\textbf{\Ours} & \textbf{0.745} & \textbf{0.0248} & \textbf{1.000} & \textbf{0.0015} & \textbf{0} & \textbf{$<$0.001} \\
\texttt{deepseek-v4-pro} & 0.715 & \nodata & 0.520 & \nodata & 0 & 0.043 \\
\texttt{deepseek-v4.1-flash} & 0.715 & \nodata & 0.560 & \nodata & 0 & 0.011 \\
\texttt{kimi-k3} & 0.680 & 0.1061 & 0.605 & 0.4808 & 0 & 0.059 \\
\texttt{glm-5.3} & 0.510 & 0.3064 & 0.985 & 0.0354 & 0 & 0.219 \\
\texttt{claude-opus-5} & 0.395 & \nodata & 0.990 & \nodata & 28 & 8.599 \\
\texttt{gpt-5.6} & 0.395 & \nodata & 0.850 & \nodata & 2 & 0.345 \\
\texttt{qwen3.8-max} & 0.395 & 0.7884 & 0.545 & 0.8794 & 0 & 0.138 \\
\texttt{claude-sonnet-5} & 0.365 & \nodata & 0.960 & \nodata & 88 & 1.763 \\
\texttt{gpt-5} & 0.365 & \nodata & 1.000 & \nodata & 120 & 2.964 \\
 
\bottomrule
\end{tabular}
\begin{tabnotes}
The three rows above the rule are not systems: they are the best accuracy any predictor can reach
on these labels, the score of answering ``yes'' every time, and the $\qL$ of a constant.
\emph{No answer} counts replies that carried no usable letter even after the budget was raised to
4096 tokens; those are reported here rather than folded into the accuracy column, because
returning nothing and answering wrongly are different events. A dash under $\qL$ means the
endpoint exposes no token probabilities at all.
\end{tabnotes}
\end{table}

\begin{figure}[tp]
\centering
\includegraphics[width=0.84\textwidth]{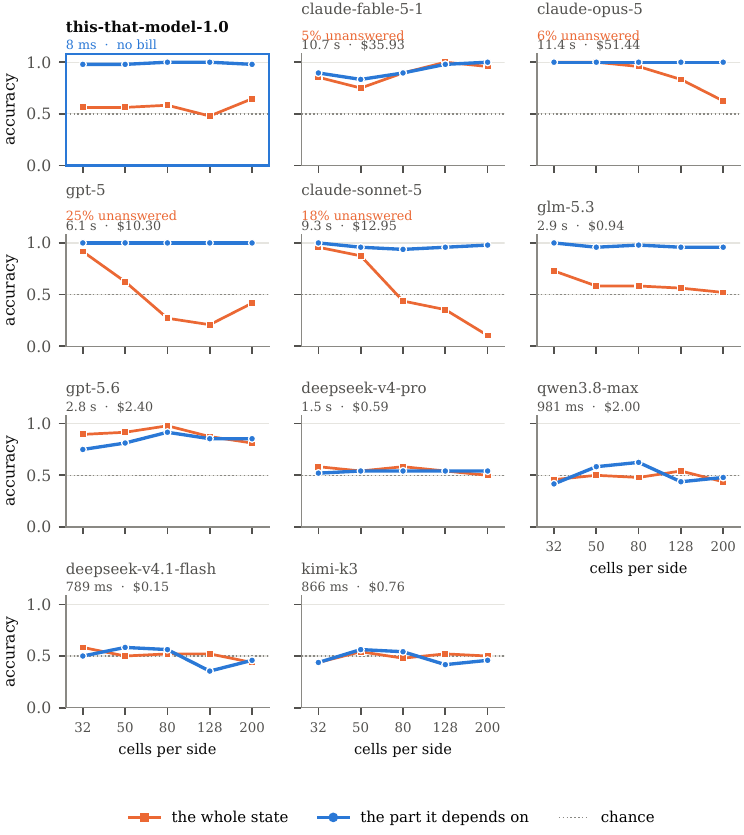}
\caption{The same ladder, one panel per system, one shared scale, ordered by where the windowed
curve ends up. Above each panel is the median time for one question and what the whole
\ladderQuestions{}-question pass billed. Accuracy is over every question \emph{asked}: a question
a system did not answer is counted as wrong, and the orange line says what share of that system's
questions those were. Scoring instead over the questions answered would condition on a subset each
system selected using the variable this figure varies: the unanswered questions sit almost
entirely on the whole-state arm and grow with the map, while the windowed arm of the same models
has none at any size.}
\label{fig:laddermodels}
\end{figure}

The decomposition is not a property of our model, and the honest way to show that is to run the
identical ladder through every hosted system we have access to (Figure~\ref{fig:laddermodels}).
The result is the same shape in every system that can do the task at all. Given the part the answer
depends on, the curve is flat and near-perfect from $32\times32$ to $200\times200$. Given the whole
map, it decays --- and the decay is not mostly wrong answers. It is silence:
\ladderSilentGlobal{}\,\% of the whole-map questions at the largest size went unanswered across the
hosted systems, against \ladderSilentLocal{}\,\% of the windowed ones, each of those failures
having spent a full \ladderBudget{}-token budget on a state of roughly \ladderStateTokens{} tokens
and returned nothing. We count those as wrong, because a decision the caller never receives is not
a decision.

The size of the effect is the part worth keeping. \ladderLiftName{} answers \ladderLiftWhole{} of
the questions correctly when handed the whole $200\times200$ map and \ladderLiftPart{} when handed
the window --- the same model, the same questions, the same worlds, differing only in how much of
the state it was given. \ladderLiftN{} of the \ladderHostedN{} hosted systems gain more than
\ladderLiftThreshold{} from the window and none loses anything by it. The cheaper systems sit near
chance under both renderings, which is what tells us the ladder is a real task rather than a
formatting exercise.

\subsection{A larger benchmark, and what it says about us}
\label{sec:spatial}

The recorded cohort of Table~\ref{tab:frozen} is \frozenN{} questions of a single shape. Our
internal suite has the opposite problem: the strongest hosted models answer most of it perfectly,
and a benchmark everyone passes ranks nothing~\cite{kiela2021dynabench,srivastava2023beyond}. So we
built a larger one and released it: \spatialTotalN{} questions over \spatialFamilies{} families and
two environments, every answer computed from the simulator rather than annotated.

Three properties are worth stating because they are what make the set usable rather than merely
large. Each family is \emph{exactly} label-balanced by construction --- items are bucketed by
answer and drawn equally from each bucket, so a constant answer scores chance everywhere, and a
family whose rare answer is genuinely rare yields fewer questions instead of being topped up. Every
item is fingerprinted by \texttt{sha256} over its rendered state, question and sorted options, and
checked against \trainFingerprints{} training questions. And the same window is rendered three ways
--- an ASCII block, the identical cells as JSON, the identical cells as prose --- with the surface
drawn at random within each family, which makes the comparison between surfaces a randomised one.

The set was built to be hard, and on its first pass it was: \Ours{} scored \spatialOursBefore{}
against a chance rate of \spatialChance{}, and was \emph{at} chance when the same maze window was
written as JSON rather than as an ASCII block. We then generated \trainQuestions{} fresh questions
from the same families --- different items, different maze windows, checked by fingerprint
\emph{and} by rendered state so that nothing in the evaluation set could appear in training --- and
trained on them. Generating supervision that is at once diverse and structurally sound is a
problem in its own right~\cite{cheng2026graphsynth,cheng2026circuitsynth,cheng2026causalsynth};
ours is the easy case of it, because the simulator that renders a state also computes the
answer, so reliability is a property of the generator rather than something to be audited
afterwards. The model now scores \spatialOursAcc{} on the full set and \spatialOursSubset{} on
the per-family subset the hosted systems were measured on (Table~\ref{tab:spatial}), a factor of
\spatialOursGain{} over where it started.

\begin{table}[t]
\centering\small
\caption{The released benchmark, \spatialN{} questions drawn equally from every family.}
\label{tab:spatial}
\begin{tabular}{l rrrr rr}
\toprule
system & tokens & accuracy & seen & unseen & no answer & cost (\$) \\
\midrule
\texttt{gpt-5.6} & 286 & 0.897 & 0.947 & 0.879 & 42 & 7.38 \\
\texttt{claude-opus-5} & 510 & 0.892 & 0.908 & 0.886 & 141 & 109.26 \\
\textbf{\Ours{}} & 0 & \textbf{0.844} & 0.873 & 0.834 & 0 & free \\
\texttt{claude-sonnet-5} & 722 & 0.836 & 0.792 & 0.853 & 278 & 29.12 \\
\texttt{Jev}$^{\dagger}$ & 0 & 0.803 & 0.717 & 0.833 & 0 & \nodata \\
\texttt{gpt-5} & 1212 & 0.800 & 0.700 & 0.837 & 365 & 28.20 \\
\texttt{glm-5.3} & 219 & 0.789 & 0.687 & 0.827 & 10 & 1.44 \\
\texttt{kimi-k3} & 7 & 0.659 & 0.750 & 0.626 & 0 & 0.52 \\
\texttt{qwen3.8-max} & 1 & 0.642 & 0.720 & 0.613 & 0 & 1.33 \\
\texttt{claude-fable-5-1} & 349 & 0.631 & 0.883 & 0.539 & 94 & 81.93 \\
\texttt{deepseek-v4.1-flash} & 89 & 0.512 & 0.605 & 0.478 & 0 & 0.21 \\
\texttt{deepseek-v4-pro} & 1 & 0.425 & 0.448 & 0.417 & 0 & 0.41 \\
 
\midrule
answering constantly & \nodata & \spatialChance{} & \nodata & \nodata & \nodata & \nodata \\
\bottomrule
\end{tabular}
\begin{tabnotes}
Every
hosted system met these fifteen question shapes for the first time at test. \Ours{} was trained on
all fifteen of them --- on different items, over different maze windows, neither of which occurs in
the evaluation set --- so its row is an in-distribution number placed beside zero-shot ones, which
is a real product fact and a weak scientific one. The honest reading is not that a
\paramCountShort{} model overtook \texttt{claude-sonnet-5}: it is that a question shape you can
generate training data for costs \$0 and \oursMsPerDecision{}\,ms per decision afterwards, and one
you cannot stays where it was. \emph{tokens} is the mean generated per question, and
\emph{no answer} is how many questions that system left unanswered, counted as wrong.
$\dagger$: \texttt{Jev} is the one row we did not measure --- we hold no key for
that endpoint and it was run for us on this same subset by a third party. It is the closest
comparison here, being the only other system that answers with no generated tokens at all. Its
\emph{seen} column is lower than its \emph{unseen} column; that split is by our training mixture
and means nothing for anyone else, so for it the difference is only that those four families are
harder.
\end{tabnotes}
\end{table}

What that training fixed, it fixed completely. The surface dependence is gone:

\par\addvspace{0.9em}
{\centering\small
\begin{tabular}{l rrr}
\toprule
the state rendered as & before & after & chance \\
\midrule
an ASCII block         & \surfAsciiBefore{} & \surfAscii{} & \surfAsciiChance{} \\
the same cells as JSON & \surfJsonBefore{}  & \surfJson{}  & \surfJsonChance{} \\
the same cells as prose & \surfRowsBefore{} & \surfRows{}  & \surfRowsChance{} \\
\bottomrule
\end{tabular}\par}
\addvspace{0.9em}

\noindent The JSON column began at chance and ends at \surfJson{}. Whatever the model had learned
before was a format; what it has now survives the format changing underneath it.

What the training did not fix is more interesting, and the benchmark localises it precisely. Of the
fifteen families, \searchFamN{} questions --- \searchFamShare{}\,\% of the set --- need a search
over the map rather than a local or relative property: \texttt{first\_move} (\famFirstMove{}),
\texttt{reachable\_within} (\famReachableWithin{}), and \texttt{distance\_band}
(\famDistanceBand{}), averaging \searchFamAcc{}. Every other family is between 0.77 and 0.998. The
entire remaining gap is there.

\paragraph{The same quantity, asked two ways.} Our four held-out families --- shapes that appear in
no training question anywhere --- separate along the same line. \texttt{same\_line}, which needs
only relative geometry, reaches \hoSameLine{} against a chance rate of \hoSameLineChance{};
\texttt{reach\_two}, local counting, reaches \hoReachTwo{}. But \texttt{distance\_parity} sits at
\hoDistanceParity{} against \hoDistanceParityChance{}, and \texttt{plan\_progress} at
\hoPlanProgress{} against \hoPlanProgressChance{}. Both of those need a shortest path.

That comparison is across different mazes, which leaves an excuse: perhaps the parity questions
landed on harder maps. So we removed it. On \probeN{} mazes we asked for the shortest-path distance
twice --- once as one of five bands, the readout the model was trained on, and once as even or odd,
a readout it has never seen. The two questions are about the same number in the same maze. Where
the band is right, the parity is right \probeCond{} of the time. A coin is $0.500$.

The model is not computing the distance and reading it out two ways. It is recognising which band a
picture belongs to, and when asked for any other function of the same quantity it has nothing to
offer. A benchmark score on \texttt{distance\_band} therefore does not mean what the family name
suggests, and we would rather say so than let the number stand unqualified.
\section{Conclusion}

A decision is not a document. When a program needs to know which of $n$ named things is the case,
the useful reply is an index and a number, and every mechanism between the model's hidden state and
that index --- the decoding loop, the token budget, the parser, the retry --- is machinery in the
way. Removing it is what this paper is about, and the effect is arithmetic rather than
architectural: \oursMsPerDecision{}\,ms per decision, \oursPerSecond{} decisions per second on one
consumer card, no generated tokens, no bill, and no state leaving the machine. That is a component
a program can put inside a loop, which a \apiclaudeopusfiveMs{}\,ms call is not.

\paragraph{Against \texttt{Jev}.} The closest comparable system is
\texttt{Jev}~\cite{almeida2026systemone}, which sells the same interface --- declared options, a distribution
over exactly those options --- as a hosted endpoint. On the only inputs on which the two can be
compared without either party choosing them, the \frozenN{} questions over \frozenStates{} states
that a third party recorded, we score \frozenOursAcc{} against its \jevAcc{}, with a Brier score of
\frozenOursBrier{} against \jevBrier{} (Table~\ref{tab:frozen}). We take that comparison seriously
and state its limits: \frozenN{} questions is a small cohort, its wording is theirs rather than
ours, and the accuracy gap of \frozenGapPP{} percentage points rests on \frozenGapN{} questions.

\texttt{Jev} was run for us on the released benchmark of Section~\ref{sec:spatial}, on the
identical \spatialN{}-question subset every hosted system answered: it scores \jevBenchAcc{}
against our \spatialOursSubset{}. That comparison is the closest like-for-like in this paper ---
the same interface, the same declared options, no generated tokens on either side --- and it is
still not a clean one, because we trained on these fifteen question shapes and it had no reason
to. What can be said without qualification is narrower: of the two systems selling this interface,
ours is the one you can retrain on your own decision shapes, and doing so moved it from
\spatialOursBefore{} to \spatialOursSubset{}.

The more useful differences are not on that table. Jev publishes \$0.042 per million input tokens
with output free, and an end-to-end response time of 70--500\,ms; we measured
\oursMsPerDecision{}\,ms end to end, below the bottom of that range, and our marginal cost is
electricity on a card already bought rather than a price that must cover serving and margin --- a
difference of about one order of magnitude, not the five that separates us from a frontier LLM, and
we say so in Figure~\ref{fig:frozen} rather than letting the axis imply otherwise. The substantive
difference is where the guarantee comes from. Both systems claim that a malformed answer cannot
occur; ours follows from Equation~\eqref{eq:head}, whose support is the declared option set, and
theirs is asserted on the same structural grounds --- their own write-up notes that the zero
type-error figure is added to its plots because schema matching is guaranteed by construction
rather than because it was measured. We agree with the reasoning. We also think a released
checkpoint, an open inference stack and a published benchmark are what let a reader check it, and
that is the form in which we are shipping this.

\paragraph{What the measurements actually support.} Three claims survive our own attempts to break
them. Decomposition is real and is not ours: on the context ladder every system that can do the
task at all is flat and near-perfect given the part the answer depends on, and decays given the
whole map, with \ladderSilentGlobal{}\,\% of the whole-map questions at the largest size returning
nothing at all (Figure~\ref{fig:laddermodels}). Calibration is real and is rare: against a computed
answer distribution we reach \simOursAccEvent{} where the ceiling is \simCeilEvent{}, while most
hosted endpoints expose no probability at all and every one that does is worse than a constant
predictor (Table~\ref{tab:calib}). And generated tokens do not buy accuracy on decisions of this
kind --- on the stochastic-actuator family the systems that generate the most are the least accurate
(Figure~\ref{fig:thinking}).

\paragraph{What does not.} The benchmark we built and released began as a list of things this
model could not do, and one round of targeted data removed most of it: \spatialOursBefore{} to
\spatialOursAcc{}, with the JSON rendering going from chance to \surfJson{}. What survived that
round is sharper for having survived it. \searchFamN{} of the questions need a search over the
map, and they sit at \searchFamAcc{} while everything else is above 0.77. Asked for the same
shortest-path distance as a band and as a parity, the model answers the band and then flips a
coin on the parity, at \probeCond{} where a coin is $0.500$ --- so the band is a label it
recognises, not a number it holds. That is what a single forward pass through a fixed stack of
layers can and cannot be made to do, and no amount of data of that shape moved it.

We would rather a reader leave with the narrow claim than the wide one. A frontier model is a better judge and a worse component; the gap
between them is paid in latency and tokens rather than in intelligence; this model will answer a
bounded question about a state in \oursMsPerDecision{}\,ms for the cost of the electricity, and
will not search a graph for you at any price. A great many branches in real software need only the
bounded question; the ones that need a search should be given a search.

\section*{Availability}

\begin{sloppypar}
\Ours{} is released on the Hugging Face Hub at
\texttt{huggingface.co/\allowbreak flock-io/\allowbreak this-\allowbreak that-\allowbreak model-1.0} under the MIT licence,
with the inference package and the scripts that reproduce every measurement in this paper at
\texttt{github.com/\allowbreak FLock-io/\allowbreak this-\allowbreak that-\allowbreak model}. The weights load into the
typed-decision stack described in Section~\ref{sec:head} --- on CUDA, on Apple Silicon or on
CPU --- so an existing caller changes a model name and nothing else. It is adapted from \texttt{decider-2b},
a publicly released 2B typed-decision checkpoint and inference stack, under Apache-2.0. The
environment simulator, the stochastic-actuator construction behind Equation~\eqref{eq:q}, and the
recorded hosted-service cohort of Table~\ref{tab:frozen} come from NanoJev under the MIT licence; we
use them as published, unmodified, and thank their authors. The benchmark of Section~\ref{sec:spatial} --- \spatialTotalN{} questions over \spatialStates{}
distinct states, with the generators, the balancing and the contamination fingerprints --- is
released on the Hugging Face Hub under the MIT licence, at
\texttt{huggingface.co/\allowbreak datasets/\allowbreak limberc/\allowbreak this-\allowbreak that-\allowbreak spatial-bench}. The internal suite is an
instrument, not a release. Correspondence:
\texttt{ai@flock.io}.
\end{sloppypar}

\bibliographystyle{plainnat}
\bibliography{refs}

\end{document}